%% file: iclr2023_conference.tex
\documentclass{article} 
\usepackage{iclr2023_conference,times}

\input{math_commands.tex}

\usepackage{hyperref}
\usepackage{url}
\usepackage{svg}
\usepackage{subcaption}
\usepackage{caption}
\newcommand{\OurModel }{{RotInvVAE }}
\title{Domain Generalization in Robust Invariant Representation}

\author{Gauri Gupta\\
Massachusetts Institute of Technology\\
\texttt{gaurii@mit.edu}\\
\And
Ritvik Kapila\\
University of California San Diego\\
\texttt{rkapila@ucsd.edu}\\
\AND
Keshav Gupta\\
Thapar Institute of Engineering and Technology\\
\texttt{keshavgupta79r@gmail.com}\\
\And
Ramesh Raskar\\
Massachusetts Institute of Technology\\
\texttt{raskar@media.mit.edu}\\
}

\iclrfinalcopy 
\begin{document}

\maketitle

\begin{abstract}

Unsupervised approaches for learning representations invariant to common transformations are used quite often for object recognition. Learning invariances makes models more robust and practical to use in real-world scenarios. Since data transformations that do not change the intrinsic properties of the object cause the majority of the complexity in recognition tasks, models that are invariant to these transformations help reduce the amount of training data required. This further increases the model's efficiency and simplifies training. In this paper, we investigate the generalization of invariant representations on out-of-distribution data and try to answer the question: Do model representations invariant to some transformations in a particular seen domain also remain invariant in previously unseen domains? Through extensive experiments\footnote{\url{https://github.com/GauriGupta19/Domain-Generalisation-in-Invariance}}, we demonstrate that the invariant model learns unstructured latent representations that are robust to distribution shifts, thus making invariance a desirable property for training in resource-constrained settings.
\end{abstract}

\section{Introduction}

In the real world, two images of the same object might only be related through some identity-preserving transformations. Many interesting data properties have these inherent symmetries but are represented in a way that does not attend to these symmetries. Prior work has revealed that incorporating these correspondences in the network can improve model performance significantly and make it more robust to variations in the data \cite{ref3}. Invariance in deep neural networks refers to a model's ability to produce the same output for a given input, regardless of certain changes in the input. For instance, when presented with an image of an object, a translation-invariant model will produce the same result regardless of the object's location in the image. The network achieves this property by detecting the presence of certain features in a local neighborhood.

Prior theoretical work shows that the complexity in recognition tasks is predominantly due to simple transformations such as rotation, translation, viewpoint, and illumination nuisances that swamp the intrinsic characteristics of the object \cite{ref19, ref20}. Making a model invariant to such transformations helps reduce the amount of training data required because the model does not have to learn to recognize objects in all possible positions and orientations \cite{ref24}.

Utilizing prior knowledge on intra-class variance resulting from transformations is an efficient technique that can be utilized in critical use cases with limited training data \cite{ref23, ref25}. Since most downstream tasks, including object recognition and label prediction, are invariant to specific group actions like translations and rotations, invariant models are extremely useful \cite{ref21, ref26}. We conjecture that the crux of object detection is to achieve invariance to identity-preserving transformations without losing discriminability.

Object recognition has potential applications across various fields, including Agriculture, where it can be used to identify and count crops, monitor crop health, and detect the presence of pests and diseases \cite{ref17}. Also, in the Healthcare domain, it can help in identifying and tracking patients, monitoring vital signs, and assisting with diagnosis and treatment \cite{ref18}. However, these technologies are generally difficult to adopt in developing countries due to either data scarcity or limited training resources. Thus, building resource-efficient object recognition systems is of utmost importance as it can help bridge the digital divide, provide access to technology and services, and improve people's lives, especially in developing countries. These systems can also help to reduce costs, increase efficiency, and create new opportunities for economic development. 

In this paper, we show how learning invariant model representations is a resource-efficient solution to the same underlying problem of object recognition. We are interested in algorithms that can generalize well on previously unseen data, as humans are capable of doing. Consider a scenario in which a model is trained on patient data from a specific population. Now, with each new patient, the model training process must be repeated. This can be time-consuming and thus problematic, especially in medical diagnostics where time is critical. Now, given a model trained on a particular domain, we explore the notion of generalization of model performance on a new unseen domain. We show that invariant representations learn domain-agnostic information from training data, which is then used to generalize the classifier to new, previously unseen data without retraining. 

Overall, the paper examines invariance to identity-preserving transformations as a property that is robust to domain shifts, i.e., invariant model generalization on new unseen data, imitating human-like recognition. We find that despite using a very simple classifier (thresholding the similarity between object representations), the model achieves strong performance in these highly unconstrained cases as well.

\section{Related Work}
Invariant networks use symmetries in the data which improves their performance and potentially reduces the amount of data required for training \cite{ref14}. Invariance can either be learned or can be explicitly embedded in the network. The former approach includes techniques like Data Augmentation which involve generating new data samples from the original data by applying various transformations such as rotation, scaling, cropping, etc \cite{ref12}. Although this improves generalization performance, it is inefficient in terms of training time and compute resources \cite{ref7}. In explicit invariant integration, the model is designed in a way that imposes constraints on the functions that are learned by the network, which therefore restricts the model's architecture design \cite{ref10}. For instance, graph neural networks have been used to establish powerful prediction models through message passing on graph-structured data that are invariant to permutation symmetries \cite{ref6}. Invariance embedding in convolutional neural networks (CNNs) has brought a paradigm shift in the analysis of images by detecting equivariant and invariant image features \cite{ref5}. Weight sharing is another approach for incorporating invariance. This involves using the same set of weights for multiple network parts, like different filters in a convolutional layer. This helps the network learn more general features that are invariant to transformations. Invariance embedding in convolutional neural networks (CNNs) has brought a paradigm shift in the analysis of images by detecting equivariant and invariant image features \cite{ref5}. Deep transformation-invariant approaches have also been used for clustering and aligning images \cite{ref16}.

While all of these previous works provide various methods for learning invariances in the network, our work focuses on introducing transfer learning in object recognition by utilizing invariance generalization in domain shift, which is perfectly useful in data-limited and resource-constrained settings.

\section{Robustness of invariant representations on out-of-distribution data}
We propose that deep models invariant to certain transformations should also generalize well to out-of-distribution data, i.e., they should generate invariant representations even on data that the model was not previously trained on. For instance, if a model is well trained on a particular dataset $X_{1}$ making its representation invariant to rotation, the model should also be invariant to the rotations in data $X_{2}$ it has never seen before. We investigate this notion of invariance for identity-preserving transformations by performing experiments to verify if the model learned a data agnostic invariance and disentanglement of information in the latent space. In particular, we examine the possibility of representations that are invariant to all the task-irrelevant variabilities present in the datasets. To the best of our knowledge, we are the first ones to investigate this claim for invariant deep learning models.

\section{Method} \label{method}
If a group $G$ has an action on a data-space $X$, i.e, $g(x) \neq x, x \in X, g \in G$. The invariant encoder $\eta$ maps the elements in the same orbit (here same class) in $X$ to the same point (orbit) $z \in Z = X/G $ $ \forall $ $ g \in G$, where $z$ is the invariant representation in the latent space of all the data points in the same orbit. That is, $\eta(x) = z$ $ \forall$ $ x \in O_x, O_x = \{ g(x)$ $|$ $ g \in G \} $. However, the decoder $\delta$ at best can map the invariant embedding $\eta(x)$ to an element in the orbit of $x$, i.e., $\delta(\eta(x)) \in O_x$ for some $g \in G$. We also need to extract the information of the group action $g \in G$ under which the element is transformed. Only then can we recover the original element in $O_x$. An encoder thus maps data points to its invariant representation $z$ and equivariant group action $g$, both of which are then used as input to the decoder to reconstruct the original object. During inference, when $g$ is identity, we get the object in the standard viewpoint. This approach is general enough to be extended to any kind of group transformation or even the composition of different transformations. For instance, \cite{ref1, ref21} show how the above approach can be extended to rotations, translations, their composition, and other general coordinate transforms.

\begin{figure}[!hbt]
   \centering
    {\includegraphics[scale =0.6]{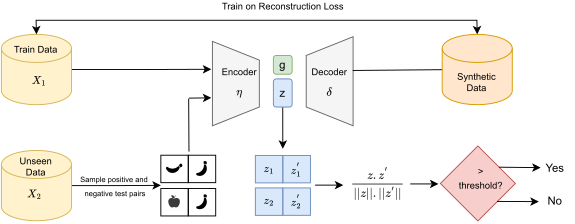}}
    \caption{\small{Evaluation framework for pairwise matching}}
    \label{figs_iclr:pipeline}
\end{figure}

In this paper, we study the problem of identification or pair-matching, e.g., for face verification. Given two images of objects never encountered during training, that are transformed under some particular transformation, the task is to decide if they depict the same object or not. We used the following procedure to assess the model's adaptability to previously unseen out-of-distribution data. The basic pipeline is shown in Figure \ref{figs_iclr:pipeline}. First, we train the invariant model on dataset $X_{1}$ transformed under some group $G$. To test the model's performance on unseen data, a classifier classifies two images of objects not seen before from dataset $X_{2}$ as either ``same" or ``different" based on a threshold. The classifier simply calculates the cosine similarity (a normalized dot product) between the latent representations of the two object images and outputs ``same" if it is more than a threshold and ``different" otherwise. We use this naïve classifier since the goal is to determine the effectiveness of these latent representations as a feature. We believe that accuracies for a majority of these tasks can be enhanced by using more advanced classifiers.

\section{Experiments}



The goal of these experiments is to explore the unconstrained notion of domain generalization of invariance under identity-preserving transformations. Here in these experiments, we study the performance of rotation invariance which can be easily extended to other common identity-preserving transformations. To test the model's generalization on unseen data, we first train a model on a particular domain $X_1$, which is then used to generate embeddings of the unseen domain $X_2$. A classifier then classifies two unseen images of objects from $X_2$ as same or not. Borrowing the notations from Section \ref{method}, a positive pair consists of an image and its rotated version i.e. $(x_1, gx_1), x_1 \in X_2, g \in G$. A negative pair consists of two randomly sampled images from $X_2$ such that they belong to different classes, i.e. $(x_1, x_2), x_1, x_2 \in X_2, $ $ s.t. $ $ O_{x_1} \neq O_{x_2}$. The pipeline is shown in Figure \ref{figs_iclr:pipeline}. 
In our setup, the test domain uses the entire orbit and never contains any of the same labels as the training domain. We train both a vanilla VAE and a rotational invariant VAE (\OurModel) inspired from \cite{ref1} and present our analysis in this section.

\begin{figure}[!hbt]
\captionsetup[subfigure]{labelformat=empty}
   \centering
    \begin{subfigure}{0.5\textwidth}
        \centering
        \caption{\hspace{13pt}MNIST Analysis}
        {\includegraphics[width=0.8\textwidth]{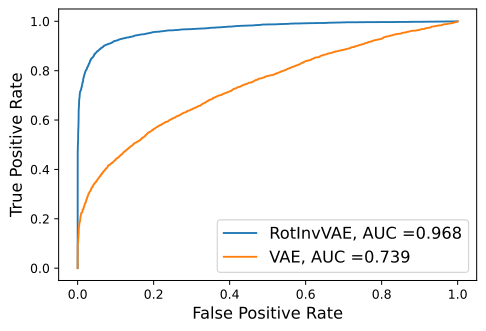}}
    \end{subfigure}%
    \begin{subfigure}{0.5\textwidth}
        \centering
        \caption{\hspace{8pt} FashionMNIST Analysis}
        {\includegraphics[width=0.8\textwidth]{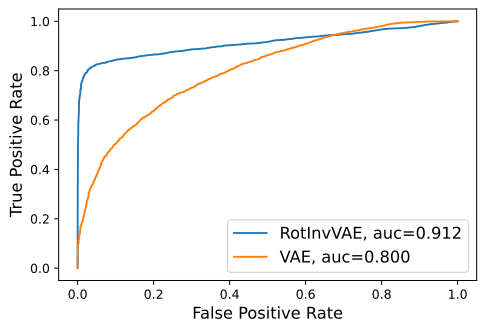}}
    \end{subfigure}
    \begin{subfigure}{0.5\textwidth}
        \centering
        {\includegraphics[width=0.8\textwidth]{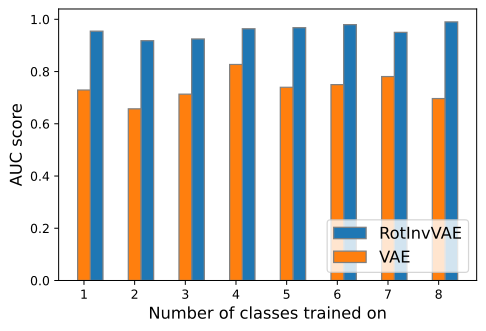}}
    \end{subfigure}%
    \begin{subfigure}{0.5\textwidth}
        \centering
        {\includegraphics[width=0.8\textwidth]{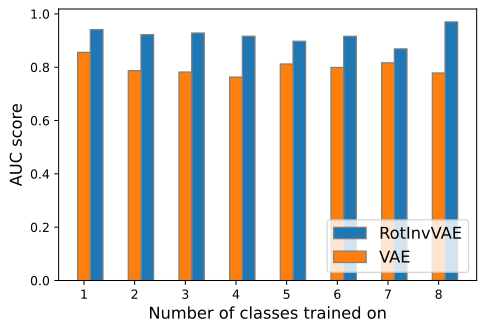}}
    \end{subfigure}
    \caption{\small{Analysis for MNIST and FashionMNIST is shown on the left and right respectively. $X_1$: Training domain, $X_2$: Testing domain (Top) ROC curves for out-of-distribution domain generalization with ${X_1}$ = rotated MNIST/FashionMNIST 0-4, ${X_2}$ = rotated MNIST/FashionMNIST 5-9 (Bottom) Area under the ROC curve (AUC) for verification task as we vary the number of the classes in ${X_1}$ where remaining classes form $X_2$}}
    \label{figs_iclr:roc_curve}
\end{figure}

We perform experiments on the following datasets: MNIST, FashionMNIST, and the Labeled Faces in the Wild (LFW) \cite{ref27}, and show ROC curves for the above-described classification task. These results clearly show \OurModel's high performance on all these datasets, depicting that the model's latent representations remain invariant to rotation even in the unseen domain. For MNIST and FashionMNIST datasets, we first train the model on the training domain $X_{1}$ which includes the labels from 0-4, and test on the remaining labels 5-9. The results can be found in Figure \ref{figs_iclr:roc_curve} (Top). Further, to verify the extent of generalization, we see that \OurModel's performance remains consistently high even if we train on fewer and fewer labels across both MNIST and FashionMNIST and test on the remaining labels as shown in Figure \ref{figs_iclr:roc_curve} (Bottom). This means that we only need to train our model on very few labels and it will still perform well on unseen labels, making the model robust to domain shift and reducing the amount of data required for training. 

\vspace{-7 pt}

\begin{figure}[!hbt]
   \centering
    \begin{subfigure}{0.5\textwidth}
        \centering
        {\includegraphics[width=0.8\textwidth]{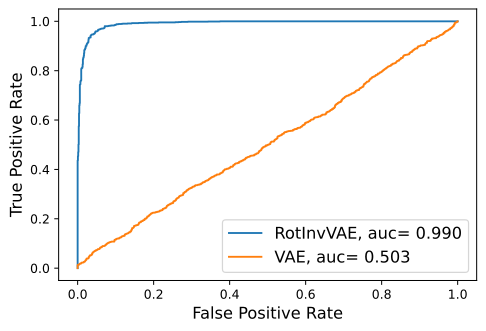}}
    \end{subfigure}%
    \begin{subfigure}{0.5\textwidth}
        \centering
        {\includegraphics[width=0.8\textwidth]{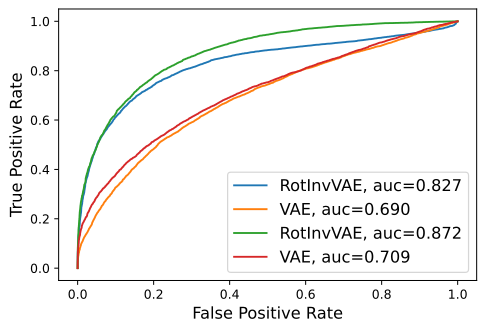}}
    \end{subfigure}
    \caption{\small{ROC curves for (Left) Face verification on unseen LFW dataset with $X_1, X_2$ described in Section \ref{impl_details} (Right) Generalization on completely different OOD data (1) blue, orange - $X_1$ = MNIST, $X_2$ = FashionMNIST (2) green, red - $X_1$ = FashionMNIST, $X_2$ = MNIST}}
    \label{figs_iclr:exp3}
\end{figure}

For investigating performance across completely unconstrained tasks, we train the \OurModel model on MNIST while testing on the FashionMNIST dataset and vice-versa, and present the consolidated results in Figure \ref{figs_iclr:exp3} (Right). It is interesting to note that the model performed well on completely different data domains, even when we train on the MNIST dataset and tested on an unseen and much more complicated domain of FashionMNIST. Additionally, the results of \OurModel on the LFW dataset face verification are even more encouraging. The implementation details and the train-test split are as described in Section \ref{impl_details}. The model essentially achieves invariance to the identity-preserving transformation (i.e., rotation in this case) and performs exceptionally well on unseen faces as well as shown in Figure \ref{figs_iclr:exp3} (Left).

\begin{figure}[!hbt]
\captionsetup[subfigure]{labelformat=empty}
   \centering
    \begin{subfigure}{1\textwidth}
        \centering   
        \caption{Vanilla VAE\hspace{150pt}\OurModel\hspace{70pt}}
        {\includegraphics[width=0.9\textwidth,height=150pt]{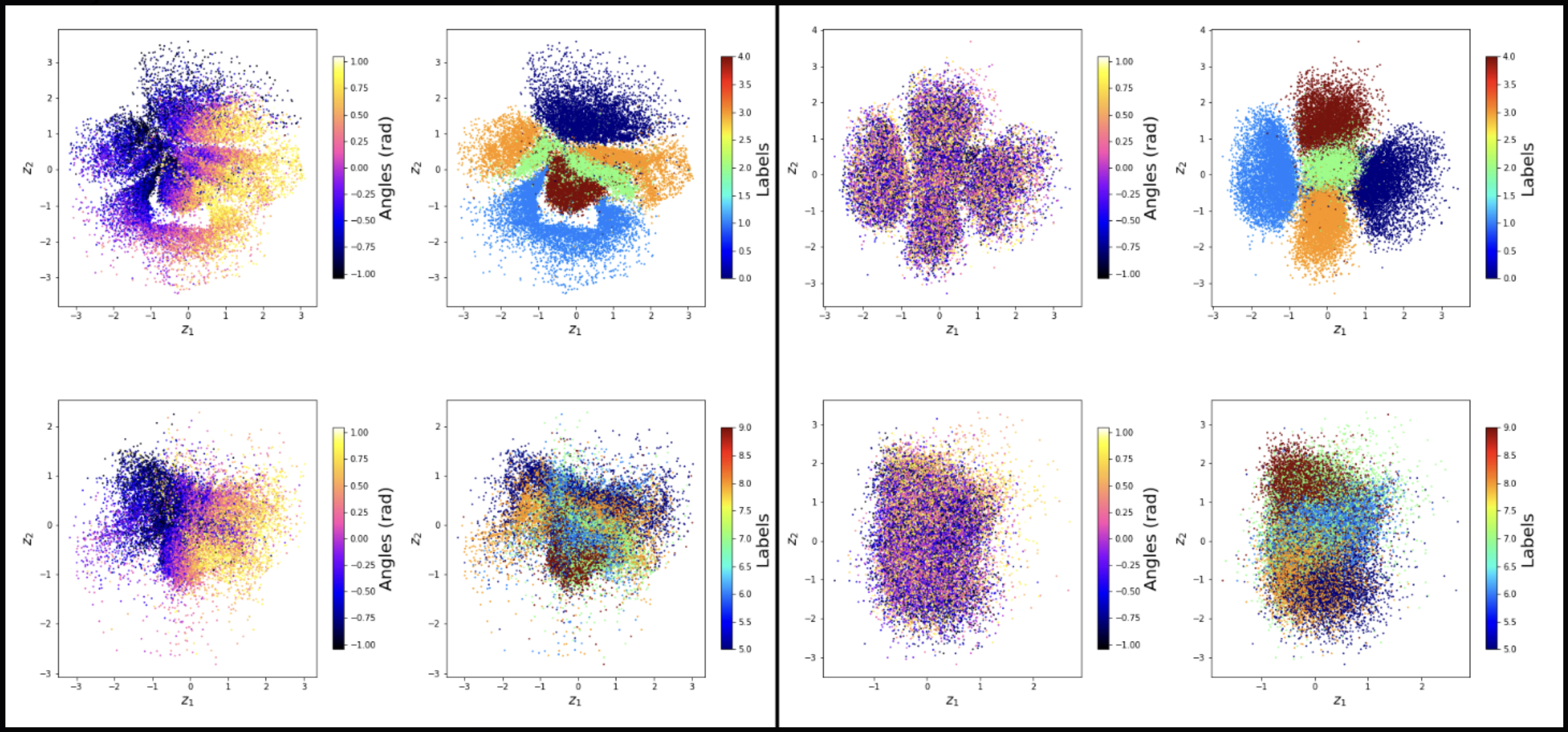}}
    \end{subfigure}%
    \caption{\small{Visualization of latent space of (Top Left) Vanilla VAE on $X_1$ = rotated MNIST 0-4 (Top Right) \OurModel on $X_1$ = rotated MNIST 0-4 (Bottom Left) Vanilla VAE on the unseen $X_2$ = rotated MNIST 5-9 (Bottom Right) \OurModel on the unseen $X_2$ = rotated MNIST 5-9}}
    \label{figs_iclr:latent_visualisation}
\end{figure}

We also visualize the latent space of both vanilla VAE and \OurModel (for dim z = 2) for the experiments in Figure \ref{figs_iclr:roc_curve} for the MNIST dataset. As in Figure \ref{figs_iclr:latent_visualisation}, we can see the latent space is divided into well-structured clusters for the training class labels. Whereas in \OurModel, as compared to vanilla VAE, the angles are randomly distributed in the latent space, indicating that the latent space is invariant to the angle of rotation for the training dataset. The latent space of \OurModel is invariant to rotations even on out-of-distribution data, which is in coherence with our hypothesis. Not only that, we also observe the emergence of clusters in the representation space for the new unseen labels for \OurModel. A similar analysis on FashionMNIST is presented in Appendix \ref{appendix}.


\subsection{Implementation details}\label{impl_details}
For implementing the LFW image dataset, we make a few transformations to make the data suitable for our task and make training faster. We first crop the images around the face and then down-sample (pixellate) them to size (50, 50). We then zero-pad the image to create a black background to perform rotation in images without inducing a bias of background effects. For the train-test split, we randomly sample face labels and include all faces with a given label in the test set until the size of the test set is one-tenth of the dataset, which gives us the set $X_{2}$. The remaining labels form our training set $X_{1}$. Images of a particular label either belong to the train or test set and are not shared across these sets. Since the goal is to test the generalization of invariant representations, it is also important to note that we compare an original image of an individual's face (which belongs to the unseen data domain) with the rotated version of the same image as a positive pair. For all these experiments, the dimension of the latent representations is consistently kept at 10 to make a fair comparison across the different models and datasets. The models are trained for 100 epochs. 

\section{Discussions and Conclusion}

In this work, we show that the model learns to generate data-agnostic representations invariant to some group transformations that also generalize well on unseen data. The model learns the group action that transforms the given data, instead of learning any particular intrinsic property of the data. Prior work only involves training and testing invariant models on the same data distribution, thus making our study unique and the first of its kind. This model property has applications in recognition classifier systems like face verification, where the input data is usually only transformed under some identity-preserving transformations. Through experiments, we show that the model generalizes well on out-of-distribution data and does not need to be retrained every time on new unseen objects. This makes the model resource and time-efficient which is particularly suitable for deployment in developing countries with limited data and training resources. Despite our recognition classifier’s simplicity, our model depicts compelling accuracy on multiple datasets. Future work could extend this approach to other objects, datasets, and additional tasks.

\bibliography{iclr2023_conference}
\bibliographystyle{iclr2023_conference}

\appendix
\section{Appendix}\label{appendix}
\begin{figure}[!hbt]
\captionsetup[subfigure]{labelformat=empty}
   \centering
    \begin{subfigure}{1\textwidth}
        \centering
        \caption{Vanilla VAE\hspace{160pt}\OurModel\hspace{70pt}}
        {\includegraphics[width=0.9\textwidth,height=150pt]{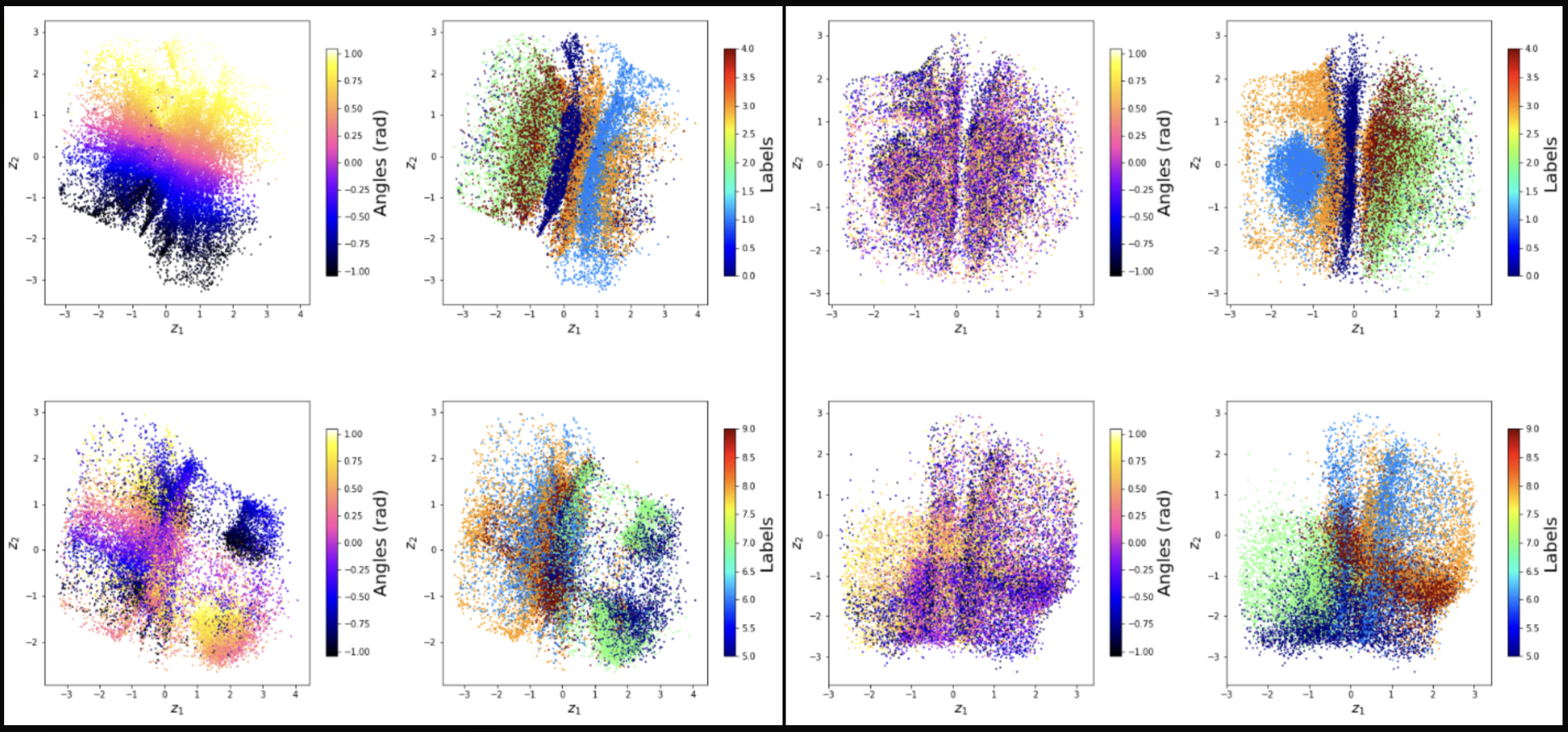}}
    \end{subfigure}
    \caption{\small{Visualization of latent space of (Top Left) Vanilla VAE on $X_1$ = rotated FashionMNIST 0-4 (Top Right) \OurModel on $X_1$ = rotated FashionMNIST 0-4 (Bottom Left) Vanilla VAE on the unseen $X_2$ = rotated FashionMNIST 5-9 (Bottom Right) \OurModel on the unseen $X_2$ = rotated FashionMNIST 5-9}}
    \label{figs_iclr:latent_visualisation_fashionMNIST}
\end{figure}


\end{document}

%% file: math_commands.tex
\usepackage{amsmath,amsfonts,bm}

\def\eqref#1{equation~\ref{#1}}

\def\1{\bm{1}}

\DeclareMathAlphabet{\mathsfit}{\encodingdefault}{\sfdefault}{m}{sl}
\SetMathAlphabet{\mathsfit}{bold}{\encodingdefault}{\sfdefault}{bx}{n}













%% file: iclr2023_conference.bbl
\begin{thebibliography}{19}
\providecommand{\natexlab}[1]{#1}
\providecommand{\url}[1]{\texttt{#1}}
\expandafter\ifx\csname urlstyle\endcsname\relax
  \providecommand{\doi}[1]{doi: #1}\else
  \providecommand{\doi}{doi: \begingroup \urlstyle{rm}\Url}\fi

\bibitem[Bepler et~al.(2019)Bepler, Zhong, Kelley, Brignole, and Berger]{ref1}
Tristan Bepler, Ellen~D. Zhong, Kotaro Kelley, Edward~J. Brignole, and Bonnie
  Berger.
\newblock Explicitly disentangling image content from translation and rotation
  with spatial-vae.
\newblock In \emph{Neural Information Processing Systems}, 2019.

\bibitem[Cohen et~al.(2019)Cohen, Weiler, Kicanaoglu, and Welling]{ref3}
Taco Cohen, Maurice Weiler, Berkay Kicanaoglu, and Max Welling.
\newblock Gauge equivariant convolutional networks and the icosahedral {CNN}.
\newblock In Kamalika Chaudhuri and Ruslan Salakhutdinov (eds.),
  \emph{Proceedings of the 36th International Conference on Machine Learning},
  volume~97 of \emph{Proceedings of Machine Learning Research}, pp.\
  1321--1330. PMLR, 09--15 Jun 2019.
\newblock URL \url{https://proceedings.mlr.press/v97/cohen19d.html}.

\bibitem[Cohen \& Welling(2016)Cohen and Welling]{ref14}
Taco~S. Cohen and Max Welling.
\newblock Group equivariant convolutional networks.
\newblock 2016.
\newblock \doi{10.48550/ARXIV.1602.07576}.
\newblock URL \url{https://arxiv.org/abs/1602.07576}.

\bibitem[Elakkiya et~al.(2022)Elakkiya, Subramaniyaswamy, Vijayakumar, and
  Mahanti]{ref18}
R.~Elakkiya, V.~Subramaniyaswamy, V.~Vijayakumar, and Aniket Mahanti.
\newblock Cervical cancer diagnostics healthcare system using hybrid object
  detection adversarial networks.
\newblock \emph{IEEE Journal of Biomedical and Health Informatics}, 26\penalty0
  (4):\penalty0 1464--1471, 2022.
\newblock \doi{10.1109/JBHI.2021.3094311}.

\bibitem[Gilmer et~al.(2017)Gilmer, Schoenholz, Riley, Vinyals, and Dahl]{ref6}
Justin Gilmer, Samuel~S. Schoenholz, Patrick~F. Riley, Oriol Vinyals, and
  George~E. Dahl.
\newblock Neural message passing for quantum chemistry.
\newblock In Doina Precup and Yee~Whye Teh (eds.), \emph{Proceedings of the
  34th International Conference on Machine Learning}, volume~70 of
  \emph{Proceedings of Machine Learning Research}, pp.\  1263--1272. PMLR,
  06--11 Aug 2017.
\newblock URL \url{https://proceedings.mlr.press/v70/gilmer17a.html}.

\bibitem[Huang et~al.(2007)Huang, Ramesh, Berg, and Learned-Miller]{ref27}
Gary~B. Huang, Manu Ramesh, Tamara Berg, and Erik Learned-Miller.
\newblock Labeled faces in the wild: A database for studying face recognition
  in unconstrained environments.
\newblock Technical Report 07-49, University of Massachusetts, Amherst, October
  2007.

\bibitem[Lecun \& Bengio(1995)Lecun and Bengio]{ref5}
Yann Lecun and Yoshua Bengio.
\newblock Convolutional networks for images, speech, and time-series.
\newblock In M.A. Arbib (ed.), \emph{The handbook of brain theory and neural
  networks}. MIT Press, 1995.

\bibitem[Lee \& Soatto(2011)Lee and Soatto]{ref19}
Taehee Lee and Stefano Soatto.
\newblock Video-based descriptors for object recognition.
\newblock \emph{Image and Vision Computing}, 29\penalty0 (10):\penalty0
  639--652, 2011.
\newblock ISSN 0262-8856.
\newblock \doi{https://doi.org/10.1016/j.imavis.2011.08.003}.
\newblock URL
  \url{https://www.sciencedirect.com/science/article/pii/S0262885611000709}.

\bibitem[Li \& Li(2021)Li and Li]{ref25}
Aoxue Li and Zhenguo Li.
\newblock Transformation invariant few-shot object detection.
\newblock In \emph{Proceedings of the IEEE/CVF Conference on Computer Vision
  and Pattern Recognition (CVPR)}, pp.\  3094--3102, June 2021.

\bibitem[Liao et~al.(2013)Liao, Leibo, and Poggio]{ref20}
Qianli Liao, Joel~Z Leibo, and Tomaso Poggio.
\newblock Learning invariant representations and applications to face
  verification.
\newblock In C.J. Burges, L.~Bottou, M.~Welling, Z.~Ghahramani, and K.Q.
  Weinberger (eds.), \emph{Advances in Neural Information Processing Systems},
  volume~26. Curran Associates, Inc., 2013.
\newblock URL
  \url{https://proceedings.neurips.cc/paper/2013/file/ad3019b856147c17e82a5bead782d2a8-Paper.pdf}.

\bibitem[Monnier et~al.(2020)Monnier, Groueix, and Aubry]{ref16}
Tom Monnier, Thibault Groueix, and Mathieu Aubry.
\newblock {Deep Transformation-Invariant Clustering}.
\newblock In \emph{NeurIPS}, 2020.

\bibitem[Rath \& Condurache(2022)Rath and Condurache]{ref23}
Matthias Rath and Alexandru~Paul Condurache.
\newblock Improving the sample-complexity of deep classification networks with
  invariant integration.
\newblock \emph{CoRR}, abs/2202.03967, 2022.
\newblock URL \url{https://arxiv.org/abs/2202.03967}.

\bibitem[Schütt et~al.(2018)Schütt, Sauceda, Kindermans, Tkatchenko, and
  Müller]{ref10}
Kristof Schütt, Huziel~E. Sauceda, P.-J Kindermans, Alexandre Tkatchenko, and
  Klaus-Robert Müller.
\newblock Schnet – a deep learning architecture for molecules and materials.
\newblock \emph{The Journal of Chemical Physics}, 148:\penalty0 241722, 06
  2018.
\newblock \doi{10.1063/1.5019779}.

\bibitem[Sohn \& Lee(2012)Sohn and Lee]{ref26}
Kihyuk Sohn and Honglak Lee.
\newblock Learning invariant representations with local transformations.
\newblock In \emph{International Conference on Machine Learning}, 2012.

\bibitem[Thomas et~al.(2018)Thomas, Smidt, Kearnes, Yang, Li, Kohlhoff, and
  Riley]{ref7}
Nathaniel Thomas, Tess~E. Smidt, Steven~M. Kearnes, Lusann Yang, Li~Li, Kai
  Kohlhoff, and Patrick~F. Riley.
\newblock Tensor field networks: Rotation- and translation-equivariant neural
  networks for 3d point clouds.
\newblock \emph{ArXiv}, abs/1802.08219, 2018.

\bibitem[van Dyk \& Meng(2001)van Dyk and Meng]{ref12}
David~A. van Dyk and Xiao-Li Meng.
\newblock The art of data augmentation.
\newblock \emph{Journal of Computational and Graphical Statistics}, 10\penalty0
  (1):\penalty0 1--50, 2001.
\newblock ISSN 10618600.
\newblock URL \url{http://www.jstor.org/stable/1391021}.

\bibitem[Winter et~al.(2022)Winter, Bertolini, Le, Noé, and Clevert]{ref21}
Robin Winter, Marco Bertolini, Tuan Le, Frank Noé, and Djork-Arné Clevert.
\newblock Unsupervised learning of group invariant and equivariant
  representations, 2022.
\newblock URL \url{https://arxiv.org/abs/2202.07559}.

\bibitem[Yang et~al.(2022)Yang, Guo, Marinello, Ercisli, and Zhang]{ref17}
Jiachen Yang, Xiaolan Guo, Francesco Marinello, Sezai Ercisli, and Zhuo Zhang.
\newblock A survey of few-shot learning in smart agriculture: developments,
  applications, and challenges.
\newblock \emph{Plant Methods}, 18, 03 2022.
\newblock \doi{10.1186/s13007-022-00866-2}.

\bibitem[Zhu et~al.(2021)Zhu, An, and Huang]{ref24}
Sicheng Zhu, Bang An, and Furong Huang.
\newblock Understanding the generalization benefit of model invariance from a
  data perspective.
\newblock \emph{CoRR}, abs/2111.05529, 2021.
\newblock URL \url{https://arxiv.org/abs/2111.05529}.

\end{thebibliography}
